\documentclass[11pt]{article}

\usepackage[margin=1in]{geometry}
\usepackage[numbers,sort&compress]{natbib}

\usepackage{microtype}

\usepackage{amsmath,amsfonts,amssymb}
\usepackage{graphicx}
\graphicspath{{Figures/}}
\usepackage{subfigure}
\usepackage{booktabs}
\usepackage{caption}
\usepackage{multirow}
\usepackage{wrapfig}
\usepackage{float}
\usepackage{xspace}
\usepackage{comment}

\usepackage[inline]{enumitem}
\setlist[enumerate]{itemsep=0pt, topsep=0pt, leftmargin=*}

\usepackage{algorithm}
\usepackage{algorithmic}

\usepackage[dvipsnames]{xcolor}
\usepackage{hyperref}
\usepackage[capitalise]{cleveref}

\usepackage[compact]{titlesec}
\newcommand{\todo}[1]{}
\newcommand{\TD}[1]{}
\newcommand{\MH}[1]{}
\newcommand{\TED}[1]{}

\renewcommand{\max}{\mathrm{max}}

\newcommand{\diag}{\mathrm{diag}}
\newcommand{\softmax}{\mathrm{softmax}}
\newcommand{\rowmax}{\mathrm{rowmax}}
\newcommand{\rowsum}{\mathrm{rowsum}}
\newcommand{\dsoftmax}{\mathrm{dsoftmax}}

\newcommand{\vQ}{\mathbf{Q}}
\newcommand{\vK}{\mathbf{K}}
\newcommand{\vV}{\mathbf{V}}

\newcommand{\vdQ}{\mathbf{dQ}}
\newcommand{\vdK}{\mathbf{dK}}
\newcommand{\vdV}{\mathbf{dV}}
\newcommand{\vS}{\mathbf{S}}
\newcommand{\vdS}{\mathbf{dS}}
\newcommand{\vP}{\mathbf{P}}
\newcommand{\vdP}{\mathbf{dP}}

\newcommand{\vO}{\mathbf{O}}
\newcommand{\vdO}{\mathbf{dO}}

\newcommand{\fa}{\textsc{FlashAttention}\xspace}
\newcommand{\faa}{\textsc{FlashAttention-2}\xspace}
\newcommand{\fat}{\textsc{FlashAttention-3}\xspace}
\newcommand{\faf}{\textsc{FlashAttention-4}\xspace}

\begin{document}

\title{\faf: Algorithm and Kernel Pipelining Co-Design for Asymmetric Hardware Scaling}

\author{%
  \begin{tabular}{c}
    Ted Zadouri\textsuperscript{*1,6} \quad
    Markus Hoehnerbach\textsuperscript{*2} \quad
    Jay Shah\textsuperscript{*3} \\
    Timmy Liu\textsuperscript{4} \quad
    Vijay Thakkar\textsuperscript{2,5} \quad
    Tri Dao\textsuperscript{1,6}
  \end{tabular}
  \\
  \makebox[\linewidth][c]{\small
    \textsuperscript{1}Princeton University \quad
    \textsuperscript{2}Meta \quad
    \textsuperscript{3}Colfax Research \quad
    \textsuperscript{4}NVIDIA \quad
    \textsuperscript{5}Georgia Tech \quad
    \textsuperscript{6}Together AI
  }
}
\date{}

\begingroup
\renewcommand{\thefootnote}{\fnsymbol{footnote}}
\maketitle
\footnotetext[1]{Equal contribution}
\endgroup

\begin{abstract}
Attention, as a core layer of the ubiquitous Transformer architecture, is the bottleneck for large language models and long-context applications.
While \fat optimized attention for Hopper GPUs through asynchronous execution and warp specialization, it primarily targets the H100 architecture.
The AI industry has rapidly transitioned to deploying Blackwell-based systems such as the B200 and GB200, which exhibit fundamentally different performance characteristics due to asymmetric hardware scaling: tensor core throughput doubles while other functional units (shared memory bandwidth, exponential units) scale more slowly or remain unchanged.
We develop several techniques to address these shifting bottlenecks on Blackwell GPUs: (1) redesigned pipelines that exploit fully asynchronous MMA operations and larger tile sizes, (2) software-emulated exponential and conditional softmax rescaling that reduces non-matmul operations, and (3) leveraging tensor memory and the 2-CTA MMA mode to reduce shared memory traffic and atomic adds in the backward pass.
We demonstrate that our method, \faf, achieves up to 1.3$\times$ speedup over cuDNN 9.13 and 2.7$\times$ over Triton on B200 GPUs with BF16, reaching up to 1613 TFLOPs/s (71\% utilization).
Beyond algorithmic innovations, we implement \faf entirely in CuTe-DSL embedded in Python, achieving 20-30$\times$ faster compile times compared to traditional C++ template-based approaches while maintaining full expressivity.

\end{abstract}

\section{Introduction}
\label{sec:intro}

The Transformer architecture~\citep{vaswani2017attention} continues to serve as the primary backbone for nearly all AI applications, from large language models~\citep{brown2020language} to vision~\citep{dosovitskiy2020image} and multimodal systems. For Transformers, the attention mechanism constitutes the primary computational bottleneck, with self-attention scores computed between queries and keys exhibiting quadratic scaling in sequence length. Scaling attention to longer contexts unlocks new capabilities such as reasoning over multiple documents~\citep{guo2021longt5,shaham2022scrolls}, modeling entire codebases~\citep{roziere2023code}, and processing high-resolution videos~\citep{chen2022scaling,ho2022video}. Meanwhile, accelerator hardware continues to evolve rapidly~\citep{nvidia2024nvidia}, with each generation delivering substantially higher peak compute throughput. However, this evolution is asymmetric: while matrix multiplication units scale aggressively, other functional units such as memory bandwidth and specialized compute units scale more slowly, creating increasingly unbalanced hardware pipelines that demand careful algorithmic co-design.

This has generated sustained interest in making attention faster through algorithmic innovations that deeply integrate knowledge of GPU hardware characteristics. \citet{dao2022flashattention} introduced \fa, which eliminates intermediate reads/writes to slow global memory through novel tiling and kernel fusion. \citet{dao2023flashattention2} restructured this as \faa to parallelize over the sequence length dimension, improving GPU occupancy. \citet{shah2024flashattention3} further adapted the algorithm for Hopper GPUs as \fat, exploiting asynchronous execution through warp specialization and incorporating FP8 support. Recent work has also explored low-precision attention: SageAttention~\citep{sageattention} achieves speedups through INT8 quantization, SageAttention2~\citep{sageattention2} extends this with INT4/FP8 quantization, and SageAttention3~\citep{sageattention3} demonstrates FP4 quantization on Blackwell consumer GPUs. However, these approaches primarily target consumer GPUs, while most AI compute is deployed on datacenter GPUs. Meanwhile, \fat primarily targets the NVIDIA Hopper H100 architecture, while the AI industry has rapidly transitioned to deploying Blackwell-based datacenter systems~\citep{nvidia2024nvidia} such as the B200 and GB200, which represent a new generation of GPUs with fundamentally different performance characteristics.

A critical trend in accelerator evolution is the asymmetric scaling of hardware units. Although Blackwell B200 doubles the tensor core throughput compared to Hopper H100 (2.25 PFLOPS vs. 1 PFLOPS for FP16/BF16), other functional units (shared memory bandwidth, exponential units, and integer/floating point ALUs) scale more slowly or remain unchanged. As a result, non-MMA resources emerge as bottlenecks. Our roofline analysis (\cref{sec:fwd,sec:bwd}) reveals that for typical attention workloads on Blackwell, surprisingly, shared memory traffic and exponential operations now dominate execution time, exceeding MMA compute by 25-60\%. Additionally, Blackwell introduces new architectural features: 256 KB of tensor memory (TMEM) per SM for storing intermediate tensor core results, 128$\times$128 MMA tiles (double the area of Hopper's 64$\times$128), and fully asynchronous tensor core operations that write directly to TMEM. Simply porting existing attention algorithms to this new hardware either leaves significant performance on the table or is impossible due to lack of forward compatibility for Hopper MMA instructions.

To this end, we propose \faf, which co-designs the algorithm and kernel implementation to address the shifting bottlenecks in modern GPU architectures. Rather than treating hardware as a uniform compute resource, we explicitly identify and mitigate bottlenecks in non-matmul units through algorithmic innovations:

\begin{enumerate}[itemsep=0pt,topsep=0pt,leftmargin=*]
\item \textbf{Redesigned pipeline for maximum overlap:} We develop new software pipelines for both forward and backward passes that exploit Blackwell's fully asynchronous MMA operations and larger tile sizes to maximize overlap between tensor cores, softmax computation, and memory operations.

\item \textbf{Exponential unit bottleneck mitigation:} For the forward pass, we implement software-emulated exponential functions using polynomial approximation on FMA units, increasing exponential throughput. We also introduce conditional softmax rescaling that skips unnecessary rescaling operations.

\item \textbf{Shared memory traffic reduction:} For the backward pass, we leverage tensor memory to store more intermediate results, reducing shared memory traffic. We also leverage Blackwell’s 2-CTA MMA mode, so each CTA stages and loads half of operand B to further reduce shared memory traffic, which we exploit to restructure the dQ step to halve the number of atomic reductions. We also implement a deterministic execution mode with minimal performance overhead, enabling reproducible training for reinforcement learning applications.

\item \textbf{Improved scheduling and resource allocation:} We develop new CTA scheduling strategies and register allocation schemes tailored to Blackwell's resource constraints and larger tile sizes.
\end{enumerate}

Beyond algorithmic innovations, we implement \faf entirely in CuTe-DSL embedded in Python, achieving 20-30$\times$ faster compile times compared to traditional C++ template-based approaches while maintaining full expressivity. This framework significantly improves developer productivity and lowers the barrier to entry, enabling researchers to rapidly prototype and deploy new attention variants without deep expertise in C++ template metaprogramming.

To empirically validate our method, we benchmark \faf on the B200 GPU and show that (1) BF16 achieves up to 1.3$\times$ speedup over cuDNN and 2.7$\times$ over the Triton implementation, (2) we achieve near-peak utilization on the shifted bottleneck resources, reaching up to $\sim$1600 TFLOPS (71\% theoretical max), and (3) for large sequence lengths, \faf outperforms alternative attention implementations.

We open source \faf with a permissive license and are working to integrate it with popular libraries to benefit the largest number of researchers and developers. The code is available at \url{https://github.com/Dao-AILab/flash-attention/tree/main/flash_attn/cute}

\section{Background}
\label{sec:background}

\subsection{Multi-Head Attention}
\label{subsec:multi_head_attn}

Let $\vQ, \vK, \vV \in \mathbb{R}^{N \times d}$ be the query, key and value input sequences associated to a single head, where $N$ is the sequence length and $d$ is the head dimension. The attention output $\vO \in \mathbb{R}^{N \times d}$ is computed as:
\begin{align*}
  \vS &= \alpha \vQ \vK^\top \in \mathbb{R}^{N \times N}, \\
  \vP &= \softmax(\vS) \in \mathbb{R}^{N \times N}, \\
  \vO &= \vP\vV \in \mathbb{R}^{N \times d},
\end{align*}
where $\softmax$ is applied row-wise and $\alpha = 1/\sqrt{d}$ is the scaling factor. In practice, we subtract $\rowmax(\vS)$ from $\vS$ for numerical stability.
For multi-head attention (MHA), each head has its own set of projections, and this computation parallelizes across multiple heads and batches.

Given output grad $\vdO \in \mathbb{R}^{N \times d}$, the backward computes:
\begin{align*}
  \vdV &= \vP^\top \vdO, \quad \vdP = \vdO \vV^\top, \\
  \vdS &= \dsoftmax (\vdP), \\
  \vdQ &= \alpha \vdS \vK, \quad \vdK = \alpha \vdS^\top \vQ,
\end{align*}
where $\dsoftmax(\vdP)$ denotes row-wise softmax gradient $\mathbf{d}s = (\diag(p) - p p^\top)\mathbf{d}p$ for $p = \softmax(s)$.

\subsection{GPU Hardware Characteristics and Execution Model}
\label{subsec:hardware}

We describe the aspects of the GPU's execution model relevant for \faf, with a focus on the NVIDIA Blackwell architecture (B200 \& GB200). We highlight key differences from the prior Hopper architecture that motivate the optimizations in \faf.

\paragraph{Memory hierarchy:}
The GPU's memories are organized as a hierarchy of data locales, with capacity inversely related to bandwidth.
Global memory (GMEM), also known as HBM, is the off-chip DRAM that is accessible to all streaming multiprocessors (SMs).
Data from GMEM are transparently cached in an on-chip L2 cache.
Next, each SM contains a small, programmer-managed, highly banked cache called shared memory (SMEM) on the chip.
Lastly, there is the register file within each SM.

Blackwell introduces a new memory level called \emph{tensor memory} (TMEM), a 256 KB on-chip memory per SM specifically designed for storing intermediate results of tensor core operations. Unlike shared memory, TMEM is warp-synchronous and tightly coupled with the tensor cores, enabling the matrix multiply-accumulate (MMA) units to write outputs directly to TMEM without consuming registers. This alleviates the extreme register pressure that plagued Hopper kernels and enables larger tile sizes. TMEM is allocated in 32-column (16 KB) granules and requires explicit programmer management for allocation, deallocation, and data movement.

\paragraph{Thread hierarchy:}
The GPU's programming model is organized around logical groupings of execution units called threads.
From the finest to the coarsest level, the thread hierarchy is comprised of threads, warps (32 threads), warpgroups (4 contiguous warps), threadblocks (i.e. cooperative thread arrays or CTAs), threadblock clusters, and grids.
Threads in the same CTA are co-scheduled on the same SM, and CTAs in the same cluster are co-scheduled on the same GPC.
SMEM is directly addressable by all threads within a CTA, whereas each thread has at most 256 registers (RMEM) private to itself.

\paragraph{Tensor cores and increased asynchrony:}
Blackwell features fifth-generation tensor cores that operate on tiles significantly larger than in previous architectures. Each MMA tensor core instruction processes $128 \times N$ tiles (typically $N =$ 128 or 256), compared to $64 \times N$ on Hopper. Crucially, Blackwell MMAs write their output directly to TMEM asynchronously, whereas Hopper MMAs write to registers. This full asynchrony enables better overlap between computation and other operations, as the MMA units no longer block on register writeback.

Hardware support for asynchrony allows for warp-specialized kernels, where the warps of a CTA are divided into producer or consumer roles that only ever issue data movement or computation \citep{warp-specialization-2011}.

\paragraph{2-CTA tensor core:}
Blackwell supports a 2-CTA tensor core MMA mode in which a CTA pair within the same thread block cluster cooperatively executes a single MMA, allowing the operation to read and write tensor memory from both CTAs. One thread in the pair initiates the MMA, but the peer CTA must be launched and remain active while the operation is in flight. Compared to single CTA MMAs that limit the dimension M to 128, the paired mode supports M = 128 or 256 by partitioning the A tile and the accumulator across the pair in dimension M and partitioning the B tile across the two CTAs in dimension N so that each CTA stages only half of B in its own shared memory while the hardware consumes the combined B tile during the multiply. This reduces redundant shared memory capacity and bandwidth, but since these operations touch tensor memory across the CTA pair, kernels must launch CTAs in fixed pairs and use a consistent 2-CTA mode for tensor memory and tensor core operations throughout the kernel.

\paragraph{Shifting bottlenecks:}
A key trend reflected in Blackwell is that tensor core throughput scales faster than other functional units. Blackwell doubles the FP16/BF16 tensor core throughput compared to Hopper (2.25 PFLOPS~\citep{nvidia2024nvidia} vs 1 PFLOPS~\citep{nvidia2022nvidia} per GPU), but shared memory bandwidth and exponential unit throughput remain unchanged or scale more slowly. This imbalance shifts the performance bottleneck away from matrix multiplication toward shared memory traffic and non-matmul operations like softmax. As our roofline analysis in \cref{sec:fwd,sec:bwd} shows, this requires careful kernel design to maximize overlap between MMA operations and these bottleneck resources.

The throughput of several hardware components on B200 (and GB200) is listed below.
\begin{enumerate}[itemsep=0pt,topsep=0pt,leftmargin=*]
  \item Tensor cores: The BF16 MMA has a throughput of 8192 ops / clock / SM, doubled from 4096 ops / clock / SM of Hopper. This can be derived from the theoretical maximum FLOPS: 2.25 PFLOPS / 1850 Mhz clock speed / 148 SMs = 8192 ops / clock / SM.
  \item Exponential unit. The multifunction unit (MUFU) on B200 and GB200 can perform 16 ops / clock / SM, the same as Hopper~\citep{cuda}. We note that B300 and GB300 GPUs have doubled the exponential throughput to 32 ops / clock / SM, though these GPUs are not yet widely available at the time of writing.
  \item SMEM: the read throughput is 128 bytes / clock / SM, the same as Hopper, as measured by microbenchmarking~\citep{luo2025dissecting}.
\end{enumerate}
We see that the MMA throughput on Blackwell has doubled compared to Hopper, but other hardware units do not necessarily get faster at the same rate. This reflects a broader trend in accelerator design: increasing the throughput of the most important components (typically matrix multiply units) to get higher performance under similar power / silicon area constraint.

\section{Algorithm}
\label{sec:algo}

\subsection{Attention forward pass}
\label{sec:fwd}

We first do a roofline analysis to show the bottlenecks of attention forward pass, which motivates our new pipeline design, as well as changes in the \fa algorithm to increase the throughput of the exponential unit and avoid most of the softmax rescaling steps.

\subsubsection{Feeds and Speeds}
\label{sec:fwd-feeds-and-speeds}

We provide intuition for our kernel design and optimizations by first analyzing the roofline, based on the throughput of the matmul units (tensor cores), shared memory (smem), and exponential unit. We note that this is a simplified analysis that does not consider all resources in the GPU (e.g., floating point math, register bandwidth, L2 bandwidth). Nevertheless, it can identify bottlenecks.

Let the shape of the tile along the length dimension of the sequence of $\vQ$ and $\vK$ be $M \times N$, and let the head dimension be $d$. We analyze the compute and memory traffic requirements to identify the performance bottleneck.

\noindent\textbf{MMA compute.}
The forward pass performs two matrix multiply-accumulate (MMA) operations per iteration: $\vQ\vK^\top$ (computing $M \times N$ output from $M \times d$ and $d \times N$ inputs) and $\vP\vV$ (computing $M \times d$ output from $M \times N$ and $N \times d$ inputs). Each MMA requires $2MNd$ floating-point operations. With a tensor core throughput of 8192 FLOPs per cycle, the total compute time is
{\setlength{\abovedisplayskip}{4pt}%
 \setlength{\belowdisplayskip}{4pt}%
 \setlength{\abovedisplayshortskip}{3pt}%
 \setlength{\belowdisplayshortskip}{3pt}%
\begin{equation}
    T_{\text{MMA}} = \frac{4MNd}{8192} \text{ cycles}.
\end{equation}}

\noindent\textbf{Shared memory traffic.}
Of the two MMAs, one is shared-shared (SS) where both operands are read from shared memory ($\vQ\vK^\top$), while the other is tensor-shared (TS) where operand $A$ is read from tensor memory and operand $B$ from shared memory ($\vP\vV$). Since each MMA instruction operates on tiles of size $128 \times 128$, computing an $M \times N$ output requires $\lceil M/128 \rceil \times \lceil N/128 \rceil$ MMA instructions. Crucially, when multiple MMA instructions are needed, the shared memory operands are read multiple times.

For $\vQ\vK^\top$ (SS), computing the $M \times N$ output requires $\lceil M/128 \rceil \times \lceil N/128 \rceil$ MMA instructions, each reading a $128 \times d$ chunk of $\vQ$ and a $d \times 128$ chunk of $\vK^\top$ from shared memory. The total shared memory reads are $\lceil M/128 \rceil \times \lceil N/128 \rceil \times (128d + 128d) = \lceil M/128 \rceil \lceil N/128 \rceil \times 256d$ elements. For $\vP\vV$ (TS), computing the $M \times d$ output requires $\lceil M/128 \rceil \times \lceil d/128 \rceil$ MMA instructions, each reading a $N \times 128$ chunk of $\vV$ from shared memory, totaling $\lceil M/128 \rceil \times \lceil d/128 \rceil \times 128N$ elements. At 2 bytes per element (bf16) and 128 bytes per cycle bandwidth, the shared memory ($T_{\text{smem}}$) read time is
{\setlength{\abovedisplayskip}{1pt}%
 \setlength{\belowdisplayskip}{0.1pt}%
 \setlength{\abovedisplayshortskip}{1pt}%
 \setlength{\belowdisplayshortskip}{0.1pt}%
\begin{equation}
  = 
  2\Big\lceil\tfrac{M}{128}\Big\rceil\Big\lceil\tfrac{N}{128}\Big\rceil 256d
+ 2\Big\lceil\tfrac{M}{128}\Big\rceil\Big\lceil\tfrac{d}{128}\Big\rceil 128N
= \tfrac{3MNd}{8192}\ \text{cycles}
\end{equation}}
(assuming $M$, $N$, $d$ are multiples of 128).

\noindent\textbf{Exponential unit.}
The exponential unit computes elementwise operations required for the softmax computation. The forward pass requires exponential operations on $M \times N$ values (corresponding to the attention matrix $\vS$). With a throughput of 16 operations per cycle, the exponential unit requires
{\setlength{\abovedisplayskip}{0.1pt}%
 \setlength{\belowdisplayskip}{1pt}%
 \setlength{\abovedisplayshortskip}{0.1pt}%
 \setlength{\belowdisplayshortskip}{1pt}%
\begin{equation}
    T_{\text{exp}} = \frac{MN}{16} \text{ cycles}.
\end{equation}}

Table~\ref{tab:fwd_roofline} summarizes the analysis for two typical tile configurations. For $M = N = d = 128$, the resources are well-balanced, with shared memory (768 cycles) being slightly lower than both MMA compute and exponential unit (both 1024 cycles). For the larger tile size $M = 256, N = d = 128$, the shared memory traffic increases to 1536 cycles due to reading MMA operands multiple times, while MMA compute and exponential unit double to 2048 cycles. This analysis motivates our kernel design to (1) use large tile sizes and maximize overlap between MMA operations and softmax computations (2) increase the throughput of exponential by using other hardware units (3) reduce the time of unnecessary non-matmul operations.

\begin{figure*}[t]
  \centering
  %\hspace*{0cm}
  \includegraphics[width=1.0\linewidth]{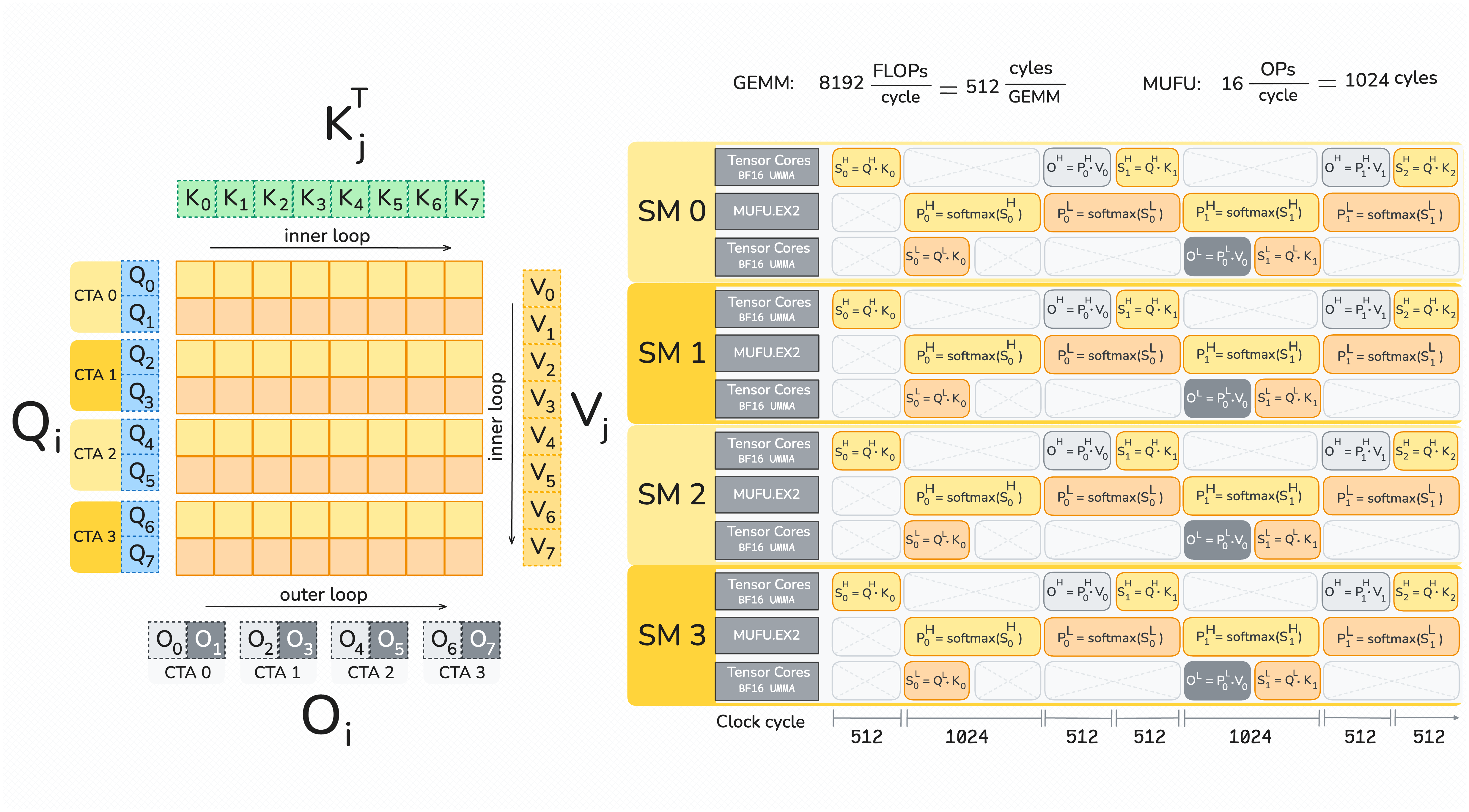}
  \caption{FlashAttention-4 forward pipeline. The superscript $^H$ denotes the matrices corresponding to the "high" Q tile, and superscript $^L$ denotes matrices corresponding to the "low" Q tile. Each Q tile corresponds to 128 query tokens.}
  \label{fig:fwd_pipeline_wide}
\end{figure*}

\begin{table}[t]
\centering
\small
\begin{tabular}{lcc}
\toprule
Resource & $128^3$ & $256 \times 128^2$ \\
\midrule
MMA compute & \textbf{1024} & \textbf{2048} \\
Shared memory & 768 & 1536 \\
Exponential unit & \textbf{1024} & \textbf{2048} \\
\bottomrule
\end{tabular}
\caption{Roofline analysis (cycles) for the attention forward pass. For both tile sizes, MMA compute and exponential unit are the primary bottlenecks.}
\label{tab:fwd_roofline}
\end{table}

\subsubsection{New pipeline to overlap matmul and softmax}

Since the Blackwell architecture doubled the tensor core flops again, taking care to overlap softmax and tensor core operations is even more crucial than on Hopper. We follow a ping-pong schedule similar to FA-3, where two tiles of the output are computed per thread block.
While one tile's tensor core operations are executed, the other tile computes softmax.
While Hopper tensor cores hold the accumulator in registers, with four threads per row in an interleaved pattern, Blackwell tensor cores hold their accumulators in tensor memory.
Additionally, a single accumulator tile on Blackwell is 128 by 128 elements large, where Hopper's tile size was 64 by 128.

The natural way to distribute work across these tiles is then to have two warpgroups of 128 threads each, with each thread processing an entire row.
This eliminates the need for inter-warp shuffles to reduce the row max, and for multiple statistics registers per thread.
Just like with FA-3, we explicitly synchronize the two softmax warpgroups to not overlap in their critical section, which is the part of exponential computation.
Each softmax warpgroup proceeds by first loading the entire row into registers, then computing the maximum, then computing the softmax (i.e., subtract the max, rescale, exponentiate, convert to input precision), and finally computing the row sum.

Another difference from FA-3 is that since we transfer $\vP$ via tensor memory rather than register file, we can decouple the rescaling of the output to a separate ``correction'' warpgroup and thus take it out of the critical path.

Several tensor memory partitionings are possible to achieve this pipeline overlap. All must allocate two tiles worth of output, leaving (at head dimension 128) half the tensor memory to store $\vS$ and $\vP$. That memory can store two copies of $\vS$ or four copies of $\vP$ (assuming the input of the FP16 or BF16 tensor core).
This leaves us with roughly two partitioning options for the remaining tensor memory: one tile of $\vS$ and two tiles of $\vP$, or two tiles of $\vS$ that overlap with $\vP$. We choose the latter because it allows us to start our software pipeline by immediately computing two $\vS$ tiles. It also leaves some tensor memory to communicate rescale statistics to the correction warpgroup.

One issue of the larger Blackwell tile sizes and the chosen thread assignment is that, unless we re-load from tensor memory, we must hold an entire row of 128 elements in register.
Given that we use two softmax warpgroups, one correction warpgroup, and one warpgroup to drive tensor cores and TMA units, assigning sufficient registers to softmax and preventing register spills is critical.
For BF16 input data types, we need to hold 128 registers for the input, and potentially 64 registers for the output (plus miscellaneous and temporary registers).
To reduce register pressure, we stage out storing $\vP$: The first three quarters are stored once (and trigger the corresponding MMA operations), and the last quarter is stored separately.

\subsubsection{Emulation of the exponential function}

\textbf{Exponential throughput bottleneck.}
On modern GPUs, the exponential function is computed by the multi-function unit (MUFU), which has significantly lower throughput than the tensor cores used for matrix multiplication. On B200 and GB200 GPUs, MUFU provides 16 operations per clock per SM, compared to 8192 operations per clock per SM for matrix multiplication. Since softmax computation requires many exponential evaluations, this disparity makes the exponential function a critical bottleneck in attention kernels.

\textbf{Software emulation via polynomial approximation.}
To increase exponential throughput, we implement a software emulation of $2^x$ using floating-point FMA units, which can operate in parallel with MUFU. We use the classical range reduction technique (Cody-Waite) and then the polynomial approximation~\citep{muller2018handbook}. The key insight is to decompose the exponential computation:
{\setlength{\abovedisplayskip}{2pt}%
 \setlength{\belowdisplayskip}{2pt}%
 \setlength{\abovedisplayshortskip}{3pt}%
 \setlength{\belowdisplayshortskip}{3pt}%
\begin{equation}
2^x = 2^{\lfloor x \rfloor}\,2^{x-\lfloor x \rfloor}
\end{equation}}
where $\lfloor x \rfloor$ is the integer part and $x - \lfloor x \rfloor \in [0, 1)$ is the fractional part.

The integer part $2^{\lfloor x \rfloor}$ can be computed efficiently using bit manipulation of the IEEE 754 floating-point representation. Since the exponent field directly represents powers of two, computing $2^{\lfloor x \rfloor}$ amounts to a shift and add operation on the exponent bits, which can be done using integer ALU instructions.

For the fractional part, we approximate $2^{x_{\text{frac}}}$ where $x_{\text{frac}} \in [0, 1)$ using a polynomial:
\begin{equation}
2^{x_{\text{frac}}} \approx \sum_{i=0}^{n} p_i\, x_{\text{frac}}^i
\end{equation}
with $p_0 = 1.0$ and the remaining coefficients chosen to minimize the relative approximation error over $[0, 1)$, calculated using the Sollya software package~\citep{chevillard2010sollya}. The polynomial evaluation uses Horner's method with FMA instructions, achieving high throughput.

The complete algorithm proceeds as follows:
\begin{enumerate}
\item Clamp $x$ to be at least $-127$ to avoid underflow
\item Compute $\lfloor x \rfloor$ using round-down mode: add $2^{23} + 2^{22}$ to $x$ (forcing the fractional bits into the mantissa), then subtract it back with round-down mode
\item Compute fractional part: $x_{\text{frac}} = x - \lfloor x \rfloor$
\item Evaluate polynomial to get $2^{x_{\text{frac}}}$
\item Combine integer and fractional parts: shift $\lfloor x \rfloor$ into the exponent field and add the mantissa bits of $2^{x_{\text{frac}}}$
\end{enumerate}

By distributing exponential computations across both MUFU and FMA units, this approach effectively increases the exponential throughput, alleviating a key bottleneck in attention computation.

\textbf{Partial emulation.}
Although polynomial emulation increases exponential throughput, it comes at a cost: additional registers (to hold intermediate values and coefficients), higher register bandwidth consumption, and longer latency compared to the MUFU instruction. Using emulation for all exponential evaluations would increase register pressure and could cause spills that negate the throughput benefit. Instead, we apply emulation to only a subset of the entries in each softmax row (10--25\%), with the remaining entries computed via hardware \texttt{MUFU.EX2}. The exact fraction is tuned empirically based on the ratio of MMA and exponential throughput for a given tile configuration.

\textbf{Numerical accuracy.}
Table~\ref{tab:ex2_accuracy} compares the accuracy of polynomial approximations of different degrees against the hardware \texttt{MUFU.EX2} instruction, measured on 4M random inputs in $[0, 1)$. We report two metrics: the FP32-level error (before any quantization) and the BF16-level error (after rounding the FP32 output to BF16), both measured against a FP64 reference.

At the FP32 level, the degree-3 polynomial has a maximum relative error of $8.8 \times 10^{-5}$, roughly $600\times$ higher than hardware. However, after rounding to BF16, the errors become nearly indistinguishable: the quantization error of BF16 (${\sim}3.9 \times 10^{-3}$) dominates the polynomial approximation error for all degrees $\geq 3$. The degree-3 polynomial matches hardware to within 1~BF16 ~ ULP on 99\% of inputs, which is sufficient for attention computation where the softmax output is consumed with BF16 precision.
Higher-degree polynomials close the FP32 gap: degree~5 matches hardware to within $2\times$ in maximum relative error, at the cost of two additional FMA instructions per evaluation.

\begin{table}[t]
\centering
\small
\begin{tabular}{lccccc}
\toprule
& \multicolumn{2}{c}{FP32 vs FP64} & \multicolumn{2}{c}{BF16 vs FP64} \\
\cmidrule(lr){2-3} \cmidrule(lr){4-5}
Method & Max rel err & Mean rel err & Max rel err & Mean rel err \\
\midrule
Ideal (FP64$\to$BF16) & --- & --- & $3.89 \times 10^{-3}$ & $1.41 \times 10^{-3}$ \\
Hardware \texttt{MUFU.EX2} & $1.41 \times 10^{-7}$ & $3.04 \times 10^{-8}$ & $3.89 \times 10^{-3}$ & $1.41 \times 10^{-3}$ \\
Degree 3 & $8.77 \times 10^{-5}$ & $5.43 \times 10^{-5}$ & $3.90 \times 10^{-3}$ & $1.41 \times 10^{-3}$ \\
Degree 4 & $3.05 \times 10^{-6}$ & $1.84 \times 10^{-6}$ & $3.89 \times 10^{-3}$ & $1.41 \times 10^{-3}$ \\
Degree 5 & $1.44 \times 10^{-7}$ & $5.48 \times 10^{-8}$ & $3.89 \times 10^{-3}$ & $1.41 \times 10^{-3}$ \\
\bottomrule
\end{tabular}
\caption{Accuracy of $2^x$ polynomial emulation on $[0, 1)$, measured against FP64 reference on 4M random inputs. FP32 columns measure the raw polynomial output; BF16 columns measure after rounding to BF16. The BF16 quantization error dominates for all degrees $\geq 3$.}
\label{tab:ex2_accuracy}
\end{table}

\subsubsection{Skipping online softmax rescaling}

\textbf{FlashAttention online softmax.}
FlashAttention computes attention $\softmax(QK^\top)V$ in blocks to minimize memory traffic. For numerical stability, the algorithm maintains running statistics as it processes blocks. When computing block $j$, let $S_j = Q K_j^\top$ be the attention scores for that block. The online softmax algorithm tracks:
\begin{align*}
m_j &= \max(m_{j-1}, \rowmax(S_j)) \\
\ell_j &= e^{m_{j-1} - m_j} \ell_{j-1} + \rowsum(e^{S_j - m_j})
\end{align*}
where $m_j$ is the running max and $\ell_j$ is the running sum of exponentials (normalizer). The intermediate output $O_j$ is updated as:
$
O_j = e^{m_{j-1} - m_j} O_{j-1} + e^{S_j - m_j} V_j.
$
The rescaling factor $e^{m_{j-1} - m_j}$ ensures numerical stability by renormalizing previous results when larger values are encountered.

\textbf{Conditional rescaling.}
The step $e^{m_{j-1} - m_j} O_{j-1}$ requires a vector multiplication. We make two simple observations:
\begin{enumerate}
\item Rescaling is only necessary when $m_j > m_{j-1}$, i.e., when new larger values are found.
\item We can tolerate some "slack" in the rescaling: only rescale when $m_j - m_{j-1} > \tau$, where $\tau$ is a threshold (typically set to $\log_2(256) = 8.0$, corresponding to a rescaling factor of 256.0). As long as we keep track of the statistics (the total scaling we have done), we can still get the true denominator at the end to get the right final output. 
\end{enumerate}

In \faf, we modify the algorithm as:
\begin{equation}
O_j = \begin{cases}
e^{m_{j-1} - m_j} O_{j-1} + e^{S_j - m_j} V_j & \text{if } m_j - m_{j-1} > \tau \\
O_{j-1} + e^{S_j - m_{j-1}} V_j & \text{otherwise}
\end{cases}
\end{equation}
When $m_j - m_{j-1} \leq \tau$, we skip updating $m$ and continue using $m_{j-1}$. This maintains the correctness because at the end of the computation, all accumulated values are renormalized by the true maximum $m_{\text{final}}$ and the final normalizer $\ell_{\text{final}}$:
{\setlength{\abovedisplayskip}{0.1pt}%
 \setlength{\belowdisplayskip}{1pt}%
 \setlength{\abovedisplayshortskip}{0.1pt}%
 \setlength{\belowdisplayshortskip}{1pt}%
\begin{equation*}
\text{Output} = \frac{1}{\ell_{\text{final}}} O_{\text{final}}
\end{equation*}}
This modification significantly reduces the number of rescaling operations while maintaining numerical accuracy, as the final normalization step corrects any small deviations introduced by skipping intermediate rescaling.

In practice, to avoid warp divergence, we rescale when any of the threads in the warp needs rescaling.

\subsection{Attention backward pass}
\label{sec:bwd}

\subsubsection{Feeds and Speeds}

Similar to the forward pass, we first provide intuition for our kernel design and optimizations by analyzing the roofline, based on the throughput of the matmul units (tensor cores), shared memory (smem), and exponential unit.

Let the tile shape along the sequence length dimension of $\vQ$ and $\vK$ be $M \times N$, and let the head dimension be $d$. We analyze the compute and memory traffic requirements to identify the performance bottleneck. Unlike for the forward pass, we make the assumption that $M = N = d = 128$ to simplify the formulas for the smem cycle count, although we retain variable names for clarity.

\noindent\textbf{MMA compute.}
The backward pass performs five matrix multiply-accumulate (MMA) operations per iteration. Each MMA involves an $M \times N$ matrix, an $M \times d$ matrix, and an $d \times N$  matrix (varying which one is the output matrix), requiring $2MNd$ floating-point operations. With a tensor core throughput of 8192 FLOPs per cycle, the total compute time is
{\setlength{\abovedisplayskip}{2pt}%
 \setlength{\belowdisplayskip}{2pt}%
 \setlength{\abovedisplayshortskip}{2pt}%
 \setlength{\belowdisplayshortskip}{2pt}%
\begin{equation}
    T_{\text{MMA}} = \frac{10MNd}{8192} \text{ cycles}.
\end{equation}}

\noindent\textbf{Shared memory traffic.}
Of the five MMAs, three of them -- $\vS^\top = \vK \vQ^\top$, $\vdP^\top = \vV \vdO^\top$, and $\vdQ = \vdS \vK$ -- are shared-shared (SS) operations where both operands are read from shared memory, while two -- $\vdV = \vP^\top \vdO$ and $\vdK = \vdS^\top \vQ$ -- are tensor-shared (TS) operations where operand $A$ is read from tensor memory and operand $B$ from shared memory. The SS MMAs read in total $2Md + 3Nd + MN$ elements from shared memory, while the TS MMAs read in total $2Md$ elements from shared memory. At a shared memory bandwidth of 128 bytes per cycle and with each element being 2 bytes (bf16), this contributes
{\setlength{\abovedisplayskip}{4pt}%
 \setlength{\belowdisplayskip}{4pt}%
 \setlength{\abovedisplayshortskip}{3pt}%
 \setlength{\belowdisplayshortskip}{3pt}%
\begin{equation}
    T_{\text{smem,MMA}} = \frac{4 M d + 3 N d + M N}{64} \text{ cycles}.
\end{equation}}
Additionally, the algorithm writes the intermediate gradient $\vdS$ of size $M \times N$ to shared memory in bf16, requiring $2MN$ bytes or $MN/64$ cycles. The gradient $\vdQ$ of size $M \times d$ is written in fp32 (4 bytes per element) to shared memory, then read back via TMA for the reduction, totaling $8Md$ bytes of shared memory traffic or $Md/16$ cycles.

The total shared memory access time ($T_{\text{smem}}$)  is therefore
{\setlength{\abovedisplayskip}{4pt}%
 \setlength{\belowdisplayskip}{4pt}%
 \setlength{\abovedisplayshortskip}{3pt}%
 \setlength{\belowdisplayshortskip}{3pt}%
\begin{equation}
    \frac{4 M d + 3 N d + M N}{64} + \frac{MN}{64} + \frac{Md}{16} \text{ cycles}.
\end{equation}}

\begin{figure*}[t]
  \centering 
  \includegraphics[width=1.0\linewidth]{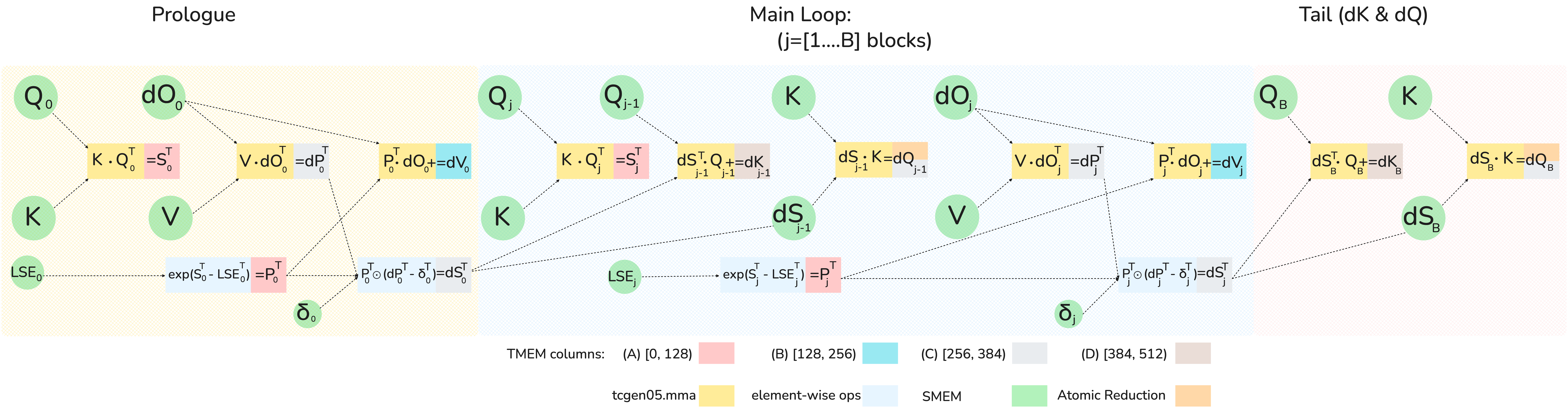}
  \caption{FlashAttention-4 backward computation graph (5 MMA operations + 2 elementwise operations), showing the 1-CTA MMA mode software pipeline order across the prologue, main loop, and tail.}
  \label{fig:bwd_computational_graph}
\end{figure*}

\noindent\textbf{Exponential unit.}
The exponential unit computes elementwise operations (exponentials, logarithms, and related nonlinear functions) required for the softmax and its gradient. The backward pass requires exponential operations on the $M \times N$ values (corresponding to the attention matrix $\vS$ and related terms). With a throughput of 16 operations per cycle, the exponential unit requires
{\setlength{\abovedisplayskip}{1pt}%
 \setlength{\belowdisplayskip}{1pt}%
 \setlength{\abovedisplayshortskip}{1pt}%
 \setlength{\belowdisplayshortskip}{1pt}%
\begin{equation}
    T_{\text{exp}} = \frac{MN}{16} \text{ cycles}.
\end{equation}}
Table~\ref{tab:bwd_roofline} summarizes the analysis for the typical tile configuration $M = N = d = 128$. The shared memory traffic time of 3328 cycles exceeds both the MMA compute time (2560 cycles) and the exponential unit time (1024 cycles), indicating that shared memory bandwidth is the primary bottleneck, though less severely than global memory traffic would suggest. This motivates our kernel design to maximize overlap between MMA operations and other computations to hide shared memory latency.

\begin{table}[t]
\centering
%\resizebox{\columnwidth}{!}{
\begin{tabular}{lcc}
\toprule
Resource & Cycles ($N = d = 128$) & Cycles ($N = d = 128$) \\
         & \textbf{1-CTA} ($M = 128$) & \textbf{2-CTA} ($M = 256$) \\
\midrule
MMA compute & 2560 & 2560 \\
Shared memory (MMA operands) & 2048 & 1536 \\
Shared memory ($\vdS$ write) & 256  & 256 \\
Shared memory ($\vdS$ \textit{DSMEM}) & 0 & 384 \\
Shared memory ($\vdQ$ write + read) & 1024 & 512 \\
%\cmidrule(lr){2-2}
\textbf{Total shared memory} & \underline{\textbf{3328}} & \underline{\textbf{2688}} \\
Exponential unit & 1024 & 1024\\
\bottomrule
\end{tabular}
%}
\caption{Roofline analysis for the attention backward pass with $M = N = d = 128$. Shared memory traffic is the bottleneck, exceeding MMA compute time by approximately 30\%. In the 2-CTA setting with $M =256$ and $N = d = 128$ (with an exception to the $\vdQ$ mma, with $M = N =128$ and $d=256$), shared memory traffic exceeds MMA compute time by approximately 5\%.}
\label{tab:bwd_roofline}
%\vspace{-20pt}
\end{table}

\subsubsection{New pipeline to overlap matmul and softmax}

The backward pass in flash attention performs five MMA operations, corresponding to recomputing $\vS$, and the two gradient computations induced by $\vQ \vK$ (which yields $\vdQ$ and $\vdK$) and $\vP \vV$ (which yields $\vdP$ and $\vdV$) respectively.
In FA-3, accumulators are stored in registers and registers are a limited resource.
This imposes significant ordering constraints that---in effect---serialize the compute graph, i.e.,  compute $\vS, \vdP, \vdV, \vdQ, \vdK$, with only the TMA load running significantly out of turn.
In addition to that, the algorithm is similar:
It iterates along the KV sequence length dimension and computes values transposed w.r.t. the forward pass, since that is the layout that the $\vdV$ and $\vdK$ gradient calculations require to read one of their operands from tensor memory.
$\vdQ$ accumulates through the atomics.

In FA-4, TMEM enables additional schedules compared to FA-3 that provide significant overlap between MMA and non-MMA operations.
Specifically, just as with the forward pass, we are trying to hide the latency of the softmax calculation.
In FA-3, the softmax computation overlaps with the MMA for $\vdP$.
From the previous section, we know that on Blackwell we need at least two MMA operations to run concurrently.

We achieve this by using the $\vdQ$ and $\vdK$ MMA's of the previous iteration.
This requires careful management of shared memory and tensor memory resources between loading, MMA, compute, and reduction operations.
In particular, note that we do not have sufficient tensor memory to fit five accumulator tiles.
At most, four tiles of 128 by 128 elements can fit and $\vdV$ and $\vdK$ accumulate, and thus cannot share their space.
In our implementation, we have $\vS$ and $\vP$ share one of the tmem blocks (at offset 0) and we have $\vdP$, $\vdS$ and $\vdQ$ share the other one. We demonstrate the computational graph of the FA-4 bwd in Figure~\ref{fig:bwd_computational_graph}.

\begin{figure*}[t]
  \centering 
  %\hspace*{-3cm}
  \includegraphics[width=1.0\linewidth]{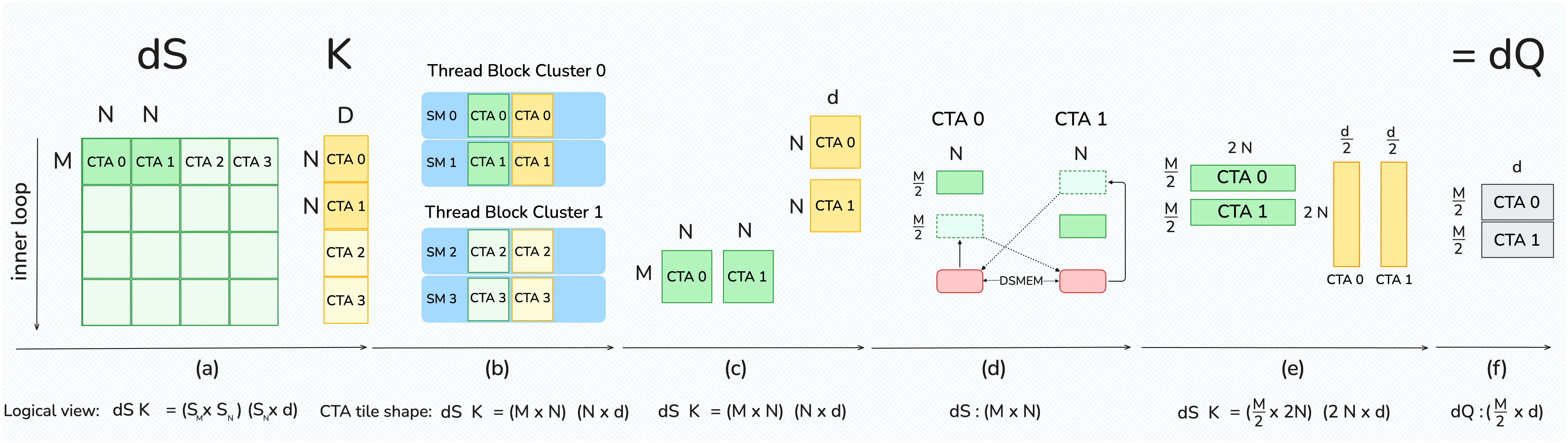}
  \caption{In the 2-CTA backward $dQ$ step, the CTA pair uses DSMEM to exchange half of the $dS$ tile so each CTA forms an $(\frac{M}{2} \times 2N)$ operand and can run a CTA-pair UMMA with a doubled reduction. 
}
  \label{fig:2cta_figure}
\end{figure*}

\subsubsection{2-CTA backward pass: Reducing shared memory traffic and global atomic adds}

Even with improved pipelining and with two of the ten GEMM operands resident in tensor memory, shared memory bandwidth still dominates the backward pass. Across the five GEMMs, the remaining eight BF16 operands are loaded from shared memory to feed the tensor cores, and this shared memory traffic incurs about 30\% more cycles than the tensor core compute. To further mitigate this bottleneck, we use the 2-CTA MMA mode introduced by Blackwell, in which the output accumulator is partitioned in the M dimension. With an MMA tile shape of $M=256$ and $N=K=128$, the two CTAs act as a single larger tile: each CTA loads and stages half of operand B and keeps only its own accumulator slice.

\textbf{Shared memory traffic}. With five GEMMs in the backward pass, we use MMA tile shape of $M=256$ and $N=K=128$, which roughly halves the shared memory traffic for operand B. In the FlashAttention backward pass, each CTA holds a fixed KV tile (parallelize the outer loop across $N$ CTAs) and streams over M tiles in the inner loop. The accumulation of $\vdQ$ is a reduction in the KV sequence in the outer loop, but the 2-CTA MMA only splits the output tile, not the reduction axis, and the reduction dimension of the $\vdQ$ MMA is $N$, which is naturally split across the CTA pair. As a result, each CTA still needs the full reduction for the rows it owns. To address this conflict on the reduction axis, we use distributed shared memory (DSMEM) to exchange half of the dS between the two CTAs since they are in the same cluster.  This approach repacks $\vdS$ so that it is partitioned along the non-reduction axis, with each CTA owning its $\frac{M}{2}$ rows and holding the full $2N$ reduction. As a result, the per CTA $\vdQ$ MMA tile shape is ($\frac{M}{2}, 2N)(2N, d)$ and accumulates a tile ($\frac{M}{2}, d$) in tensor memory. In 2-CTA MMA mode, the MMAs for $\vS$, $\vdP$, $\vdV$ and $\vdK$ run with the tile $M=256$, while $\vdQ$ uses $M=128$ but with double reduction $2N=256$.  We then reordered the software pipeline with respect to the 1-CTA variant to hide the DSMEM latency. We compute $\vdP$ for the current tile before computing $\vdQ$ for the tile of the previous iteration. The $\vdQ$ tile is small enough to fit in TMEM alongside $\vP$, reusing the same TMEM region as $\vS$, so we no longer reuse the same TMEM region for $\vdP$ and $\vdQ$ as we do in the 1-CTA mode. With this new pipelining ordering, we can compute the element-wise $\vdS$ of the current tile in parallel with the $\vdQ$ MMA of the tile from the previous iteration. See Figure~\ref{fig:2cta_figure} for an illustration of how the $\vdQ$ step is decomposed.

\textbf{$\vdQ$ atomic adds}. A complementary benefit of this $\vdQ$ decomposition is that it halves the number of global atomic reductions. Atomic updates introduce nondeterminism and are expensive as they occur in every iteration of the inner loop. Consequently, each CTA writes only half of the $\vdQ$ tile and performs half as many global atomic reductions as the 1-CTA counterpart.

\subsubsection{Deterministic backward pass} 

Our backward kernel introduces nondeterminism for the gradient computation due to inter-CTA reductions in global memory (affecting $\vdQ$ in general, and $\vdK$/$\vdV$ in the case of GQA). To ensure reproducibility and facilitate reliable debugging during training, we also provide a deterministic execution mode. The standard solution, which we also adopt, is to serialize the global reductions using a semaphore lock. Specifically, each CTA writing to a common $\vdQ$ tile must acquire the lock according to a predefined order, perform its reduction, and then release the lock by incrementing the semaphore counter. 

This lock-based approach impacts performance for two main reasons: (1) issuing the memory fence to ensure device-wide visibility of the semaphore write (required for correct acquire--release semantics), and (2) introducing stalls as each CTA waits for previous CTAs reducing on a common $\vdQ$ tile to complete. In load-imbalanced situations, a naive choice of CTA order can severely degrade performance. In general, we do CTA swizzling in the head and batch dimensions to reduce stalls (up to L2 cache capacity, cf. \cref{sec:algo_other}). For causal masking, we additionally launch KV blocks in descending order, traverse query blocks in ascending order starting from the diagonal, and order the $
\vdQ$ reductions by descending query block index. This ``shortest-processing-time-first'' (SPT) schedule ensures that no CTA is stalled on its first $\vdQ$ write.

\subsection{Scheduling}
\label{sec:algo_other}

In many situations, such as with causal masking or variable sequence length (varlen), the attention kernel is naturally load-imbalanced -- SMs are assigned worktiles whose mainloops differ in length, since some worktiles require more loads and MMAs than others. Furthermore, we can choose the order in which SMs process tiles, for instance by defining a preferred linearization of the grid coordinates. Abstracting away any specific features of attention, we can then apply general results on makespan minimization for identical parallel processors to our context. In particular, in FlashAttention-4, we use the classical idea of longest-processing-time-first (LPT) scheduling~\citep{graham1969}. We emphasize that the way in which we apply this idea works across all GPU architectures and was also validated as an improvement to FlashAttention-3 on Hopper GPUs.

\textbf{LPT for causal masking.} The standard attention grid is given by (mblocks, heads, batches) and is computed in increasing order left-to-right. But scores are masked out above the diagonal, so for fixed head and batch, SMs end up inefficiently processing worktiles from shortest to longest. On the other hand, a naive LPT order is also suboptimal, since for different batches, mainloop KV loads won't hit in L2 cache, and loading all KV heads first can thrash the L2 cache if they exceed its capacity. Instead, we always process batches as the outermost dimension, and swizzle over heads. This means that we divide heads into sections that don't overflow L2 cache; the tile scheduler then traverses the grid by heads per section, mblocks in reverse order, sections, and finally batches. In particular, for MQA or GQA, we always traverse all query heads per KV head before varying over mblocks. Empirically, we validate that this LPT order is highly effective; for example, for BF16 and head dimension 128 we obtain 4-8\% FLOPS gain for MHA and 7-14\% for MQA 8 as measured on an H200 GPU.

\textbf{LPT for variable sequence length.} For varlen, we also have to contend with load imbalance due to variation among batches. For instance, in decode workloads, different batches might attend to different amounts of context, and in mixed or continuous batching, some batches might be prefill while others are decode. The list of query and KV sequence lengths per batch is typically stored as attention metadata on device, and the standard varlen attention kernel reads these integers at runtime while processing batches in increasing order. However, the given batch order may be arbitrarily suboptimal with respect to load balancing -- for example, we could have shorter square prefills followed by long-context decodes. To ameliorate this, we can enforce an LPT order by launching a preprocessing kernel to sort batches according to their maximum per-worktile execution time, writing out the additional metadata of a virtual to actual batch index mapping that will be subsequently read back into the attention kernel in order to traverse batches in sorted order. This metadata can be cached and thus results in no performance loss from sorting.

\section{Language and Framework}
\label{sec:framework}

We write \faf entirely in CuTe-DSL~\citep{cutedsl}, embedded in Python, without any component in CUDA C++.
The CuTe-DSL compiler takes the source code in Python, lowers to PTX, then uses the PTX compiler (ptxas) to finally produce the assembly code (SASS).

\textbf{Full expressivity with clean abstractions.}
The CuTe-DSL programming model is isomorphic to CUTLASS C++, ensuring that \faf retains the full expressivity of low-level GPU programming while benefiting from the productivity gains of meta-programming in Python instead of C++ and fast JIT compilation times. CuTe-DSL provides direct access to PTX as an escape hatch, allowing developers to implement any functionality they need without framework limitations. For example, we leverage custom PTX sequences for operations not yet fully exposed in CuTe-DSL APIs (though these will be integrated in future releases), demonstrating that our framework does not constrain developers to a limited subset of GPU capabilities.

\textbf{Fast compilation through JIT.}
Compile time has been a bottleneck in past FlashAttention implementations, due to complex C++ template metaprograms. By embedding CuTe-DSL in Python with just-in-time (JIT) compilation, \faf achieves faster build times compared to traditional C++ template-based approaches. As shown in~\cref{table:compile_time}, \faf reduces compile time by 20-30$\times$ compared to \fat. This rapid iteration cycle significantly improves developer productivity, enabling faster experimentation and debugging during kernel development.

\begin{table}[h]
  \centering
  \begin{tabular}{lcc}
    \toprule
    Method & Forward pass & Backward pass \\
    \midrule
    \fat & 55s & 45s \\
    \faf & 2.5s & 1.4s \\
    \midrule
    Speedup & 22$\times$ & 32$\times$ \\
    \bottomrule
  \end{tabular}
  \caption{Compile time for a single kernel: FA3 (C++ templates) and FA4
    (CuTe-DSL). Typically FA2 and FA3 require precompiling hundreds of kernels
    for different attention variants.}
  \label{table:compile_time}
\end{table}

\textbf{Flexibility and accessibility.}
The Python based framework has already demonstrated its flexibility in practice: developers have successfully built FlexAttention and block-sparse attention variants on top of \faf without modifying the core framework. By lowering the barrier to entry, our approach enables researchers and engineers with just a few months of GPU programming experience to contribute meaningful extensions without requiring deep expertise in C++ template metaprogramming. This accessibility accelerates innovation and allows the attention mechanism research community to more rapidly explore new algorithmic variants.

Our vision is to provide a comprehensive framework for building all kinds of attention variants with best-in-class performance. Rather than implementing each attention variant from scratch, \faf factors common functionality into independent, composable primitives. Operations such as block-sparse patterns, masking strategies, variable sequence length handling, and work scheduling are all exposed as orthogonal primitives that can be freely combined. This modular design ensures that optimizations and new features benefit all attention implementations built on the framework while still achieving the highest performance by compiling down to efficient GPU kernels.

\section{Empirical Evaluation}
\label{sec:experiments}

We evaluate \faf efficiency compared to various open-source and closed-source baselines.

\paragraph{Benchmarking attention.} We measure the runtime of \faf across different sequence lengths and head dimensions, comparing it to standard implementations in PyTorch, \faa\footnote{\fat does not run on B200}, Triton (which uses B200-specific instructions~\citep{tillet2019triton}), Gluon (a lower-level GPU programming language with finer control than Triton~\citep{gluon2024}), and cuDNN (a vendor's library optimized for B200 GPUs). We confirm that \faf is up to 1.3$\times$ faster than cuDNN 9.13 and up to 2.7$\times$ faster than Triton. \faf reaches up to 1613 TFLOPs/s, approximately 71\% of the theoretical maximum TFLOPs/s on B200 GPUs.

\paragraph{Benchmark settings.} We measure runtime on a B200 GPU for different settings (with/without causal mask, head dimensions 64, 128, and (192, 128)) for BF16 inputs. We vary the sequence length as 1k, 2k, ..., 32k, and set batch size so that the total number of tokens is 32k. We set the hidden dimension to 2048, and head dimension to be either 64 or 128 (i.e., 32 heads or 16 heads). For the (192, 128) configuration used in DeepSeek V3~\citep{deepseekai2024deepseekv3technicalreport}, we use 16 heads with 192 query dimensions and 128 key/value dimensions. To calculate the FLOPs of the forward pass, we use $4 \cdot \text{seqlen}^2 \cdot \text{head dimension} \cdot \text{number of heads}$.
With causal masking, we divide this number by 2 to account for the fact that approximately only half of the entries are calculated. To get the FLOPs of the backward pass, we multiply the forward pass FLOPs by 2.5 (since there are 2 matmuls in the forward pass and 5 matmuls in the backward pass, due to recomputation).

\subsection{Forward pass}

\begin{figure*}[t]
  \centering
  \begin{minipage}{0.48\textwidth}
    \centering
    \includegraphics[width=\textwidth]{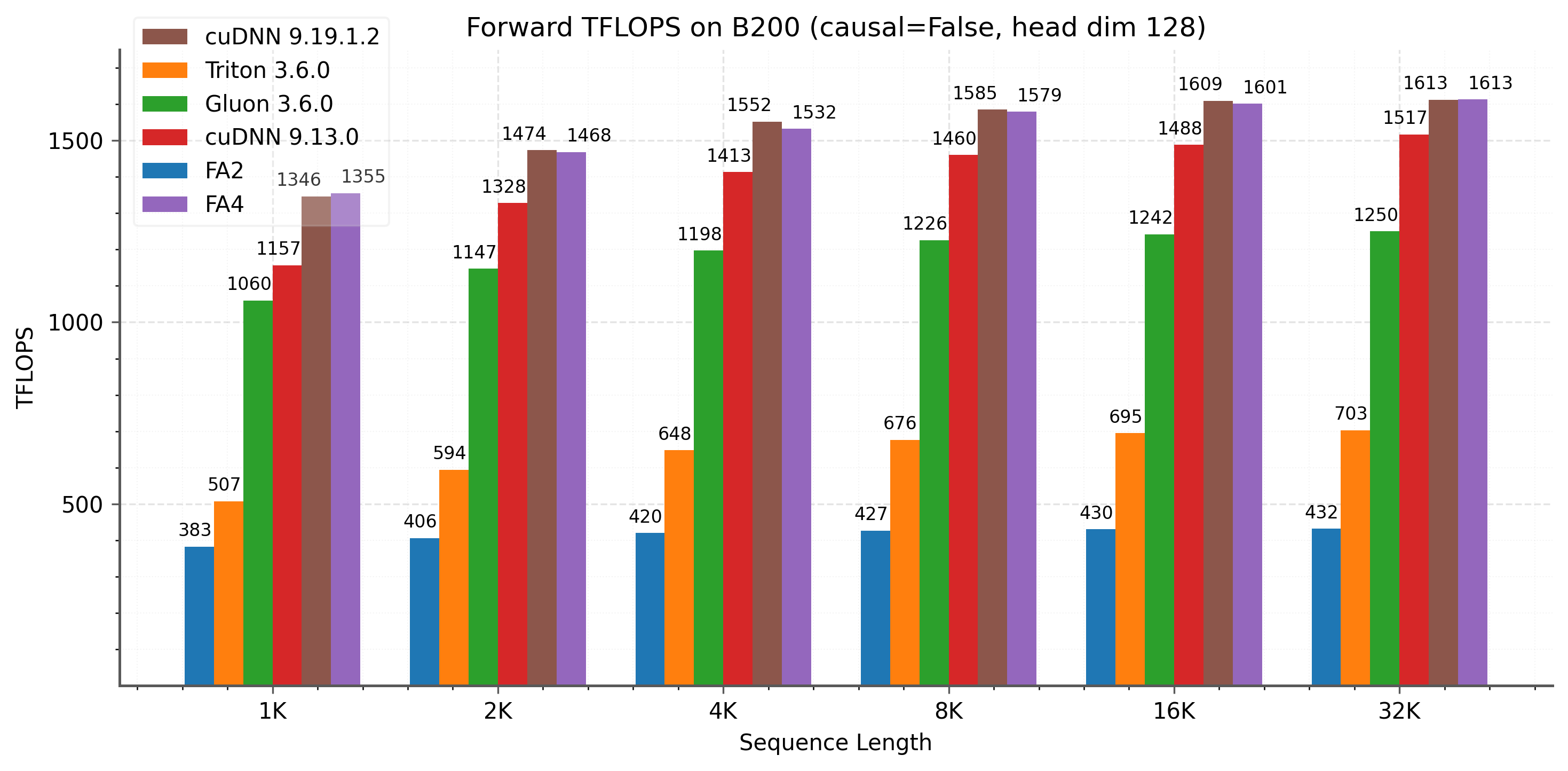}
  \end{minipage}
  \hfill
  \begin{minipage}{0.48\textwidth}
    \centering
    \includegraphics[width=\textwidth]{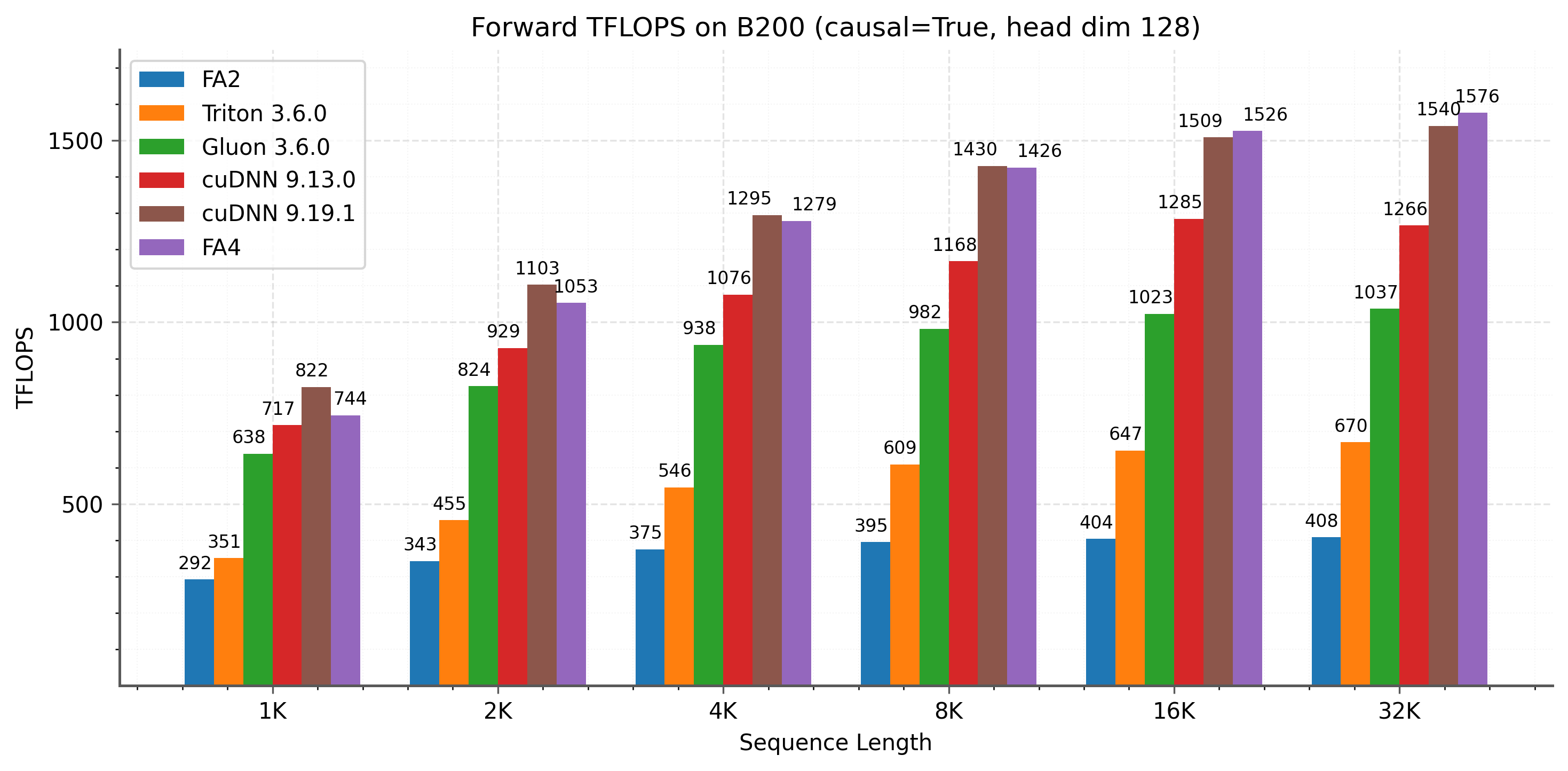}
  \end{minipage}
\caption{Forward pass TFLOPS on B200 (FP16/BF16) with head dimension 128. Left: non-causal attention. Right: causal attention. FA4 achieves 1.1-1.3$\times$ speedup over cuDNN 9.13.0 and 2.1-2.7$\times$ over Triton  across sequence lengths. Since the initial release of our implementation, newer versions of cuDNN have incorporated many of the techniques described in this paper, yielding similar performance to FA4.}
  \label{fig:fwd}
\end{figure*}

\begin{figure}[h]
  \centering
  \includegraphics[width=0.65\columnwidth]{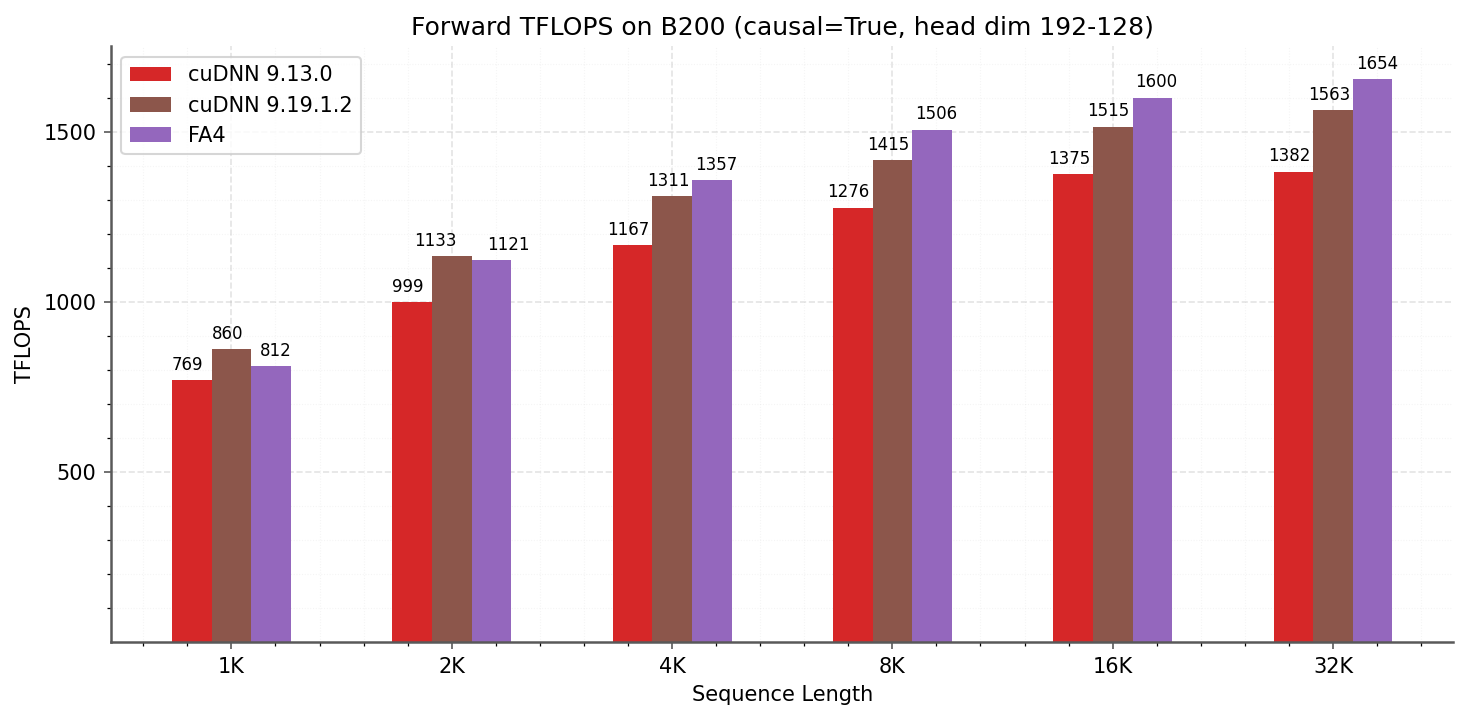}
  \caption{Forward pass TFLOPS comparison between cuDNN and FA4 on B200 (FP16/BF16) with head dimension (192, 128) for causal attention (typically used in DeepSeek V3 architecture)}
  \label{fig:fwd_hdim192128}
\end{figure}

We report forward pass results in~\cref{fig:fwd,fig:fwd_hdim192128}, showing that \faf is 1.1-1.3$\times$ faster than cuDNN 9.13 and 2.1-2.7$\times$ faster than Triton. For medium and long sequences (4k and above), \faf consistently outperforms all baselines across different head dimensions and causal masking settings.
The gains are larger for the causal case, which we attribute to the longest-processing-time first (LPT) scheduler.

\subsection{Backward pass}

\begin{figure*}[t]
  \centering
  \begin{minipage}{0.48\textwidth}
    \centering
    \includegraphics[width=\textwidth]{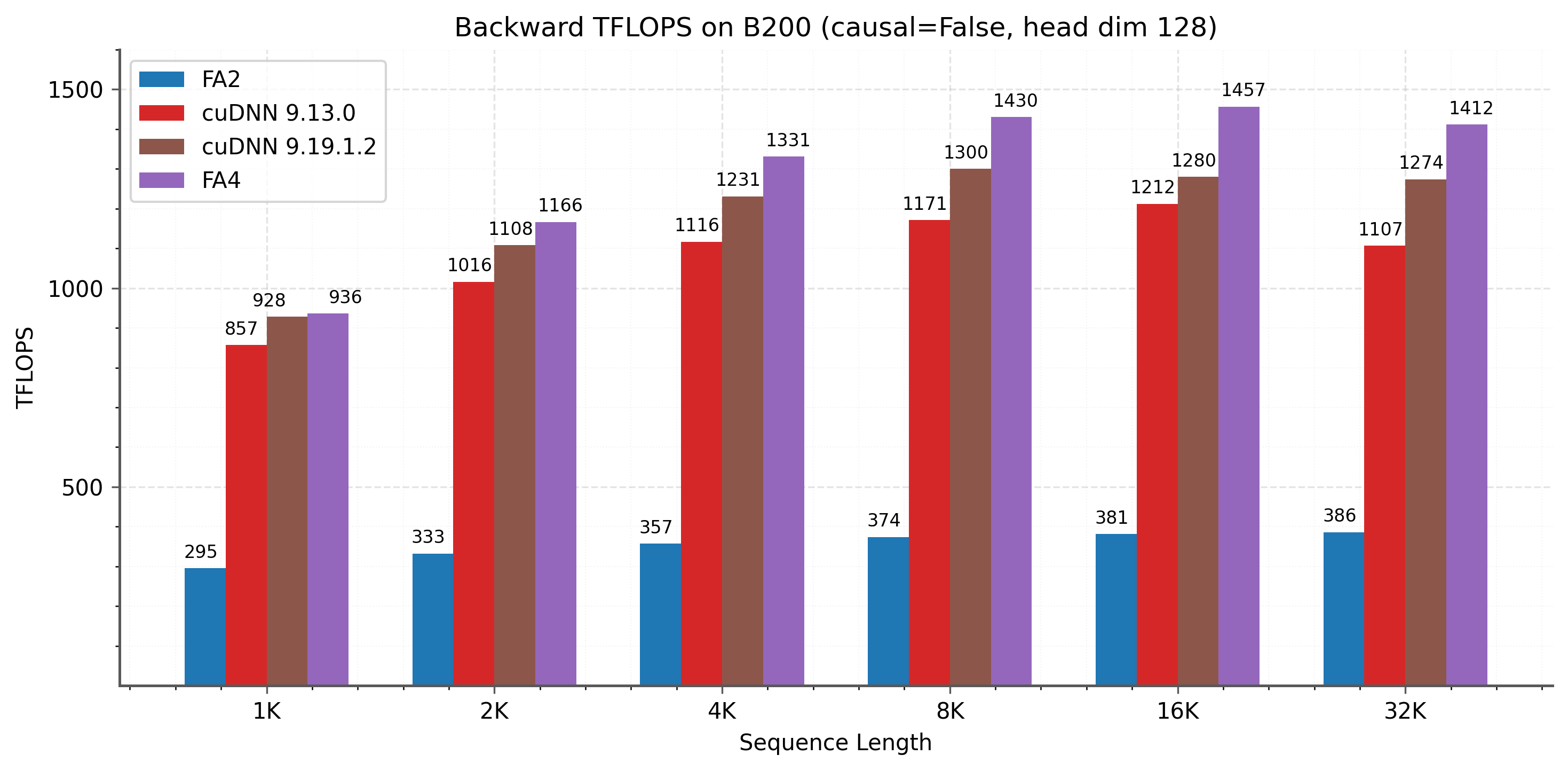}
  \end{minipage}
  \hfill
  \begin{minipage}{0.48\textwidth}
    \centering
    \includegraphics[width=\textwidth]{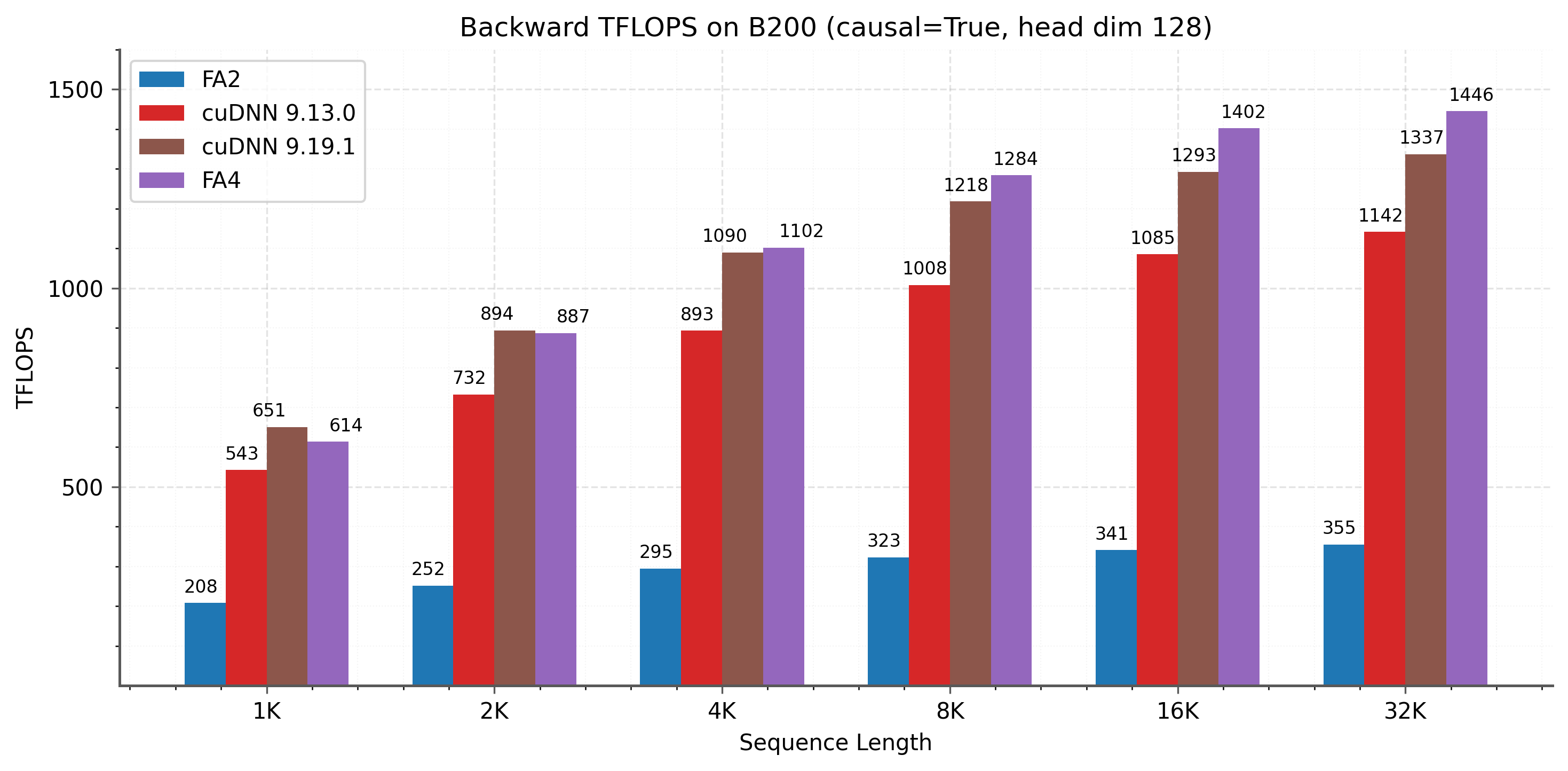}
  \end{minipage}
  \caption{Backward pass TFLOPS on B200 (FP16/BF16) with head dimension 128. Left: non-causal attention. Right: causal attention.}
  \label{fig:bwd}
\end{figure*}

We report backward pass results in~\cref{fig:bwd}. \faf achieves consistent speedups across long sequence lengths and causal masking, demonstrating the effectiveness of our 2-CTA backward pass. 

\begin{figure}[h]
  \centering
  \includegraphics[width=0.95\columnwidth]{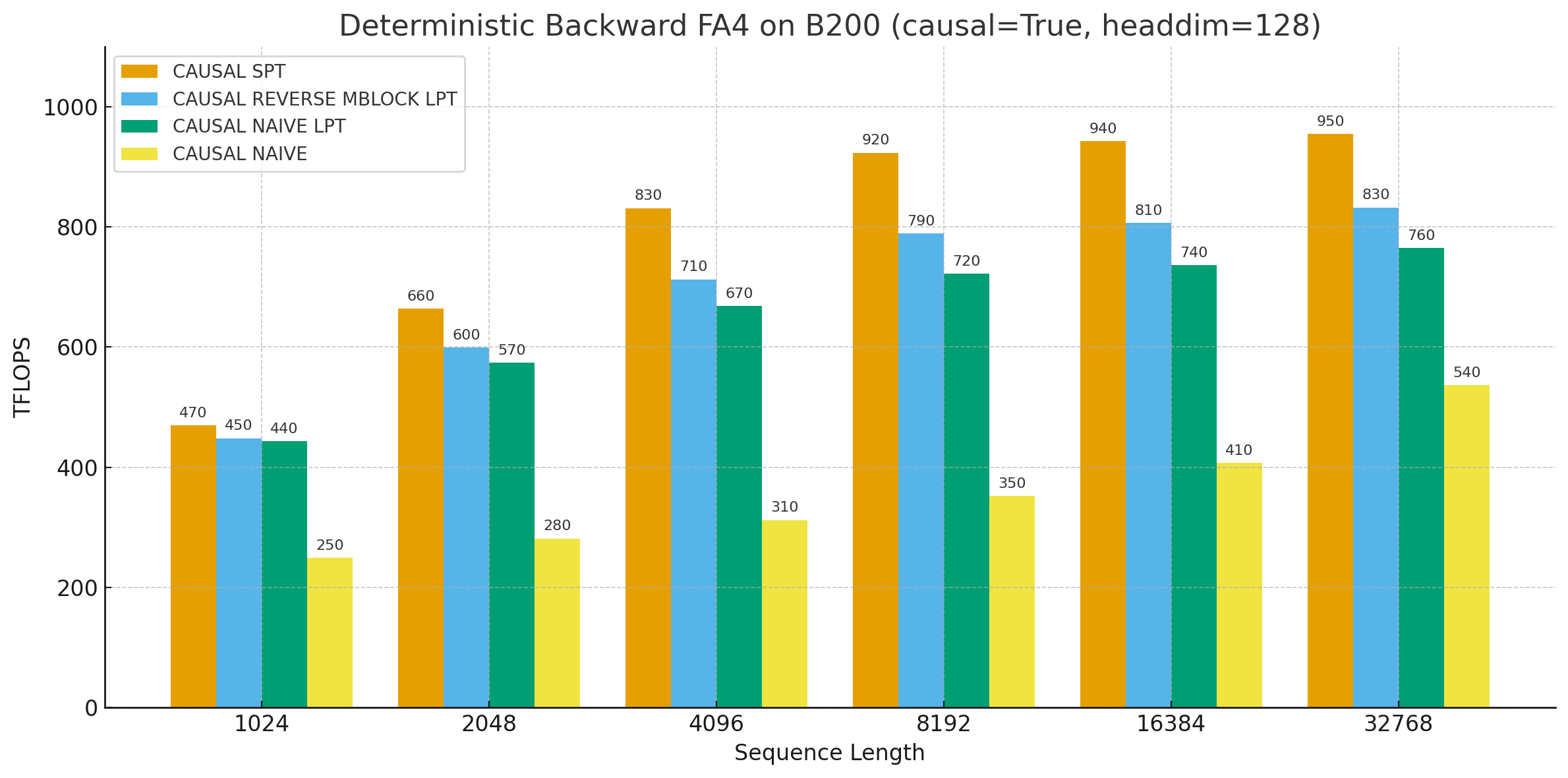}
  \caption{Ablations for Deterministic Backward pass on B200 (FP16/BF16) with head dimension 128. Causal attention -- SPT, LPT with reverse mblock order, LPT, and naive with no batch/head swizzle.}
  \label{fig:bwd-det-ablation}
\end{figure}

We also show the performance of the deterministic backward pass in~\cref{fig:bwd-det-ablation}. Our careful swizzling and scheduling results in a much faster deterministic backward pass, getting up to 75\% the speed of the nondeterministic backward pass of the 1-CTA backward pass.

\section{Discussion and Conclusion}
\label{sec:conclusion}

FlashAttention-4 addresses asymmetric hardware scaling, where tensor cores are so fast that the dominant bottlenecks shift to shared-memory traffic and exponential throughput, motivating algorithmic and kernel co-design to mitigate these limits. We redesign the pipeline around fully asynchronous MMA to overlap softmax with larger-tiled matmuls and introduce software-emulated exponential and conditional softmax rescaling to reduce non-matmul operations. We leverage tensor memory and 2-CTA MMA mode to reduce shared memory traffic. In addition, 2-CTA enables restructuring the global atomic accumulation, halving the number of global atomic adds. FlashAttention-4 is implemented entirely in CuTe-DSL embedded in Python, preserving low-level control while achieving 20-30× faster compile times than C++ template-based kernels. Although optimized for Blackwell GPUs, some of these algorithms can be extended to other accelerators as compute continues to outpace non-matmul units.

\section*{Acknowledgments}
We thank Together AI, Meta, xAI, and Princeton Language and Intelligence (PLI) for compute support. We gratefully acknowledge the support of the Schmidt Sciences AI2050 fellowship, the Google ML and Systems Junior Faculty Awards, and the Google Research Scholar program. We want to further thank the following teams at Nvidia: CuDNN, TensorRT-LLM, and Cutlass teams for constant discussions, ideas, and feedback.

\bibliographystyle{plainnat}
\bibliography{ref}

\appendix
\newpage

\section{Additional Details on Experiments and Benchmarking}

\subsection{System and libraries}
\label{sec:system}

We benchmark the speed on a B100 180GB SXM6 (1000W).
We warmup with 5 runs, then repeat the benchmarks 10 times, and take the average timing.

We generally used the latest versions of the libraries at the time of writing
(March 2025).
Specifically, we use:
\begin{itemize}
\item CUDA 13.1
\item \fa 2.8.3
\item Triton 3.6
\item PyTorch 2.10.0
\item CuTe-DSL 4.4.1 
\end{itemize}
For cuDNN, in the main paper, we compare to cuDNN 9.13 and the latest version cuDNN 9.19.1.2. Starting from version 9.13 and 9.14~\citep{cudnn913releasenotes}, we have worked with the cuDNN team to incorporate some techniques from \faf into cuDNN so that our work can benefit as many practitioners as possible.

\subsection{Backward Deterministic non-causal}

\begin{figure*}[t]
\centering
 \begin{minipage}{0.48\textwidth}
    \centering
    \includegraphics[width=\textwidth]{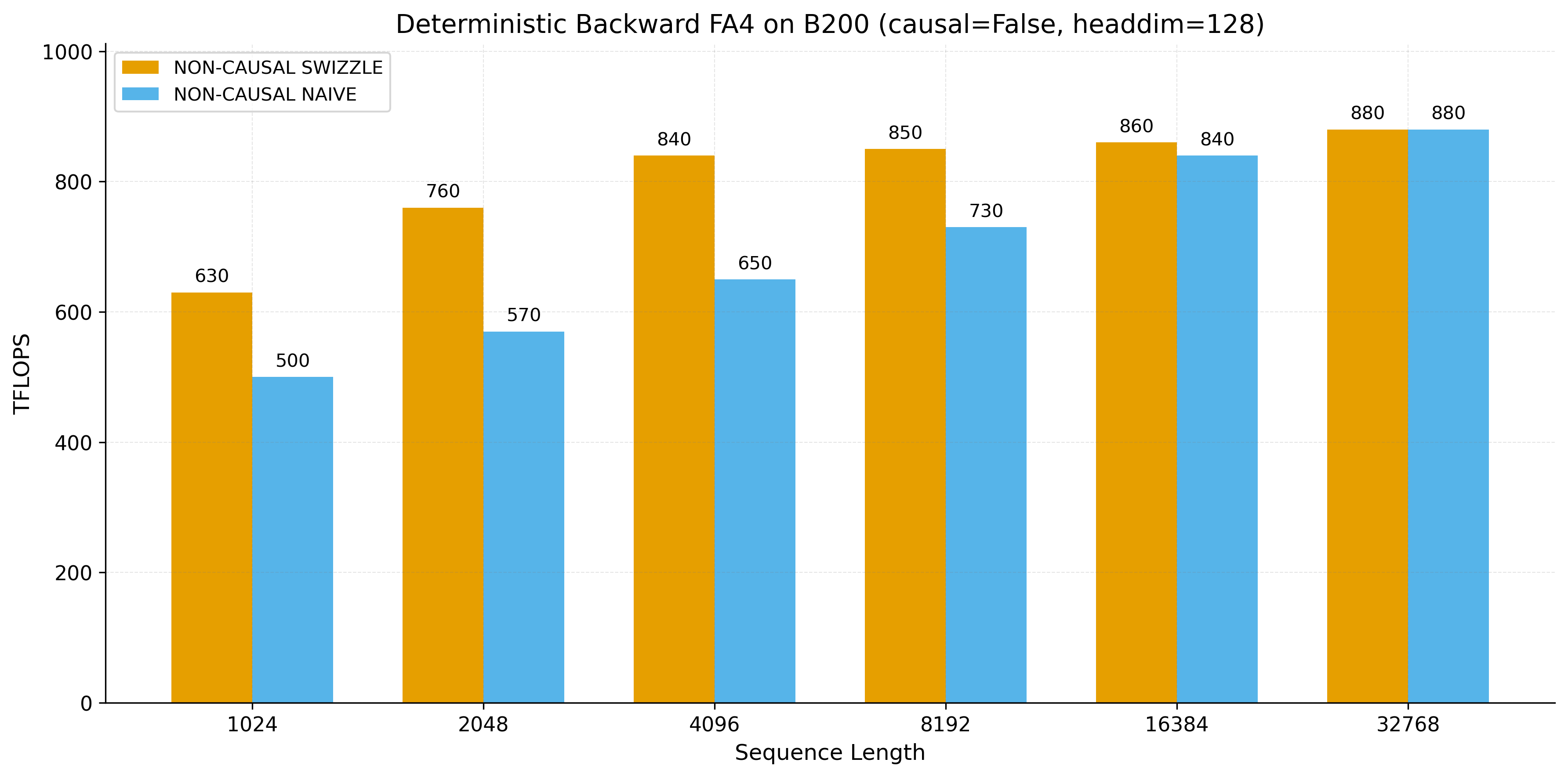}
  \end{minipage}
  \hfill
  \begin{minipage}{0.48\textwidth}
    \centering
    \includegraphics[width=\textwidth]{../Figures/causal_bwd_det_FA4.png}
  \end{minipage}
  \caption{Ablations for Deterministic Backward pass on B200 with head dimension 128. Non-causal attention with batch/head swizzle versus naive. Right: causal attention -- SPT, LPT with reverse mblock order, LPT, and naive with no batch/head swizzle.}
  \label{fig:bwd-det-ablation-non-causal}
\end{figure*}

For completeness, we also include performance numbers for the deterministic backward kernel without causal masking in \cref{fig:bwd-det-ablation-non-causal}, side by side with causal masking.

\end{document}